\documentclass[letterpaper, 10 pt, conference]{ieeeconf}  

\IEEEoverridecommandlockouts                              

\usepackage{mathtools}
\usepackage{hyperref}
\usepackage{bm} 
\usepackage{amssymb}
\usepackage{amsfonts}
\usepackage{mathbbol} 
\usepackage{graphicx, import} 
\usepackage{float}
\usepackage{times}
\usepackage{multicol}
\usepackage{multirow}    
\usepackage{microtype}
\usepackage[all]{nowidow}
\usepackage{cases}
\usepackage{multirow}
\usepackage{yfonts}
\usepackage{diagbox}
\usepackage{subcaption}
\usepackage{colortbl}
\usepackage{xparse}
\usepackage{arydshln}
\usepackage{microtype}
\usepackage{soul}
\usepackage{xcolor}
\usepackage[mediumspace,mediumqspace,squaren]{SIunits}

\usepackage{tabularx,booktabs}

\title{\LARGE \bf
Codesign of Humanoid Robots for Ergonomic Collaboration with Multiple Humans via Genetic Algorithms and Nonlinear Optimization 
}
\author{Carlotta Sartore$^{1,2}$, Lorenzo Rapetti$^{1}$, Fabio Bergonti$^{1,2}$, Stefano Dafarra$^{1}$,\\ Silvio Traversaro$^{1}$,  and Daniele Pucci$^{1,2}$
\thanks{*The code to reproduce paper results is publically available at \url{https://github.com/ami-iit/paper_sartore_2023_humanoids_codesign-ga-nl}}%
\thanks{*The paper was supported by the Italian National Institute for Insurance against Accidents at Work (INAIL) ergoCub Project.}
\thanks{$^{1}$Artificial and Mechanical Intelligence  at Istituto Italiano di Tecnologia,
Center for Robotics Technologies, Genova, Italy.
        {\tt\small name.surname@iit.it}}%
\thanks{$^{2}$School of Computer Science, The University of Manchester,
Manchester, United Kingdom.}%
}
\begin{document}

\maketitle
\thispagestyle{empty}
\pagestyle{empty}


\begin{abstract}
\emph{Ergonomics} is a key factor to consider when designing control architectures for effective physical collaborations between humans and humanoid robots. 
In contrast, \emph{ergonomic indexes} are often overlooked in the robot design phase, which leads to suboptimal performance in physical human-robot interaction tasks. 
This paper proposes a novel methodology for optimizing the design of humanoid robots with respect to ergonomic indicators associated with the interaction of multiple agents. 
Our approach leverages a dynamic and kinematic parameterization of the robot link and motor specifications to seek for optimal robot designs using a bilevel optimization approach. 
Specifically, a genetic algorithm first generates robot designs by selecting the link and motor characteristics. Then, we use nonlinear optimization to evaluate interaction ergonomy indexes during collaborative payload lifting with different humans and weights.
To assess the effectiveness of our approach, we compare the optimal design obtained using bilevel optimization against the design obtained using nonlinear optimization. Our results show that the proposed approach significantly improves ergonomics in terms of energy expenditure calculated in two reference scenarios involving static and dynamic robot motions.
We plan to apply our methodology to drive the design of the ergoCub2 robot, a humanoid intended for optimal physical collaboration with humans in diverse environments.
\looseness=-1
\end{abstract}
\section{Introduction}
\label{sec:introduction}
\begin{figure}[!t]
    \includegraphics[trim=0.0cm 0.0cm 0.0cm 0.0cm, clip=true, width=0.98\columnwidth]{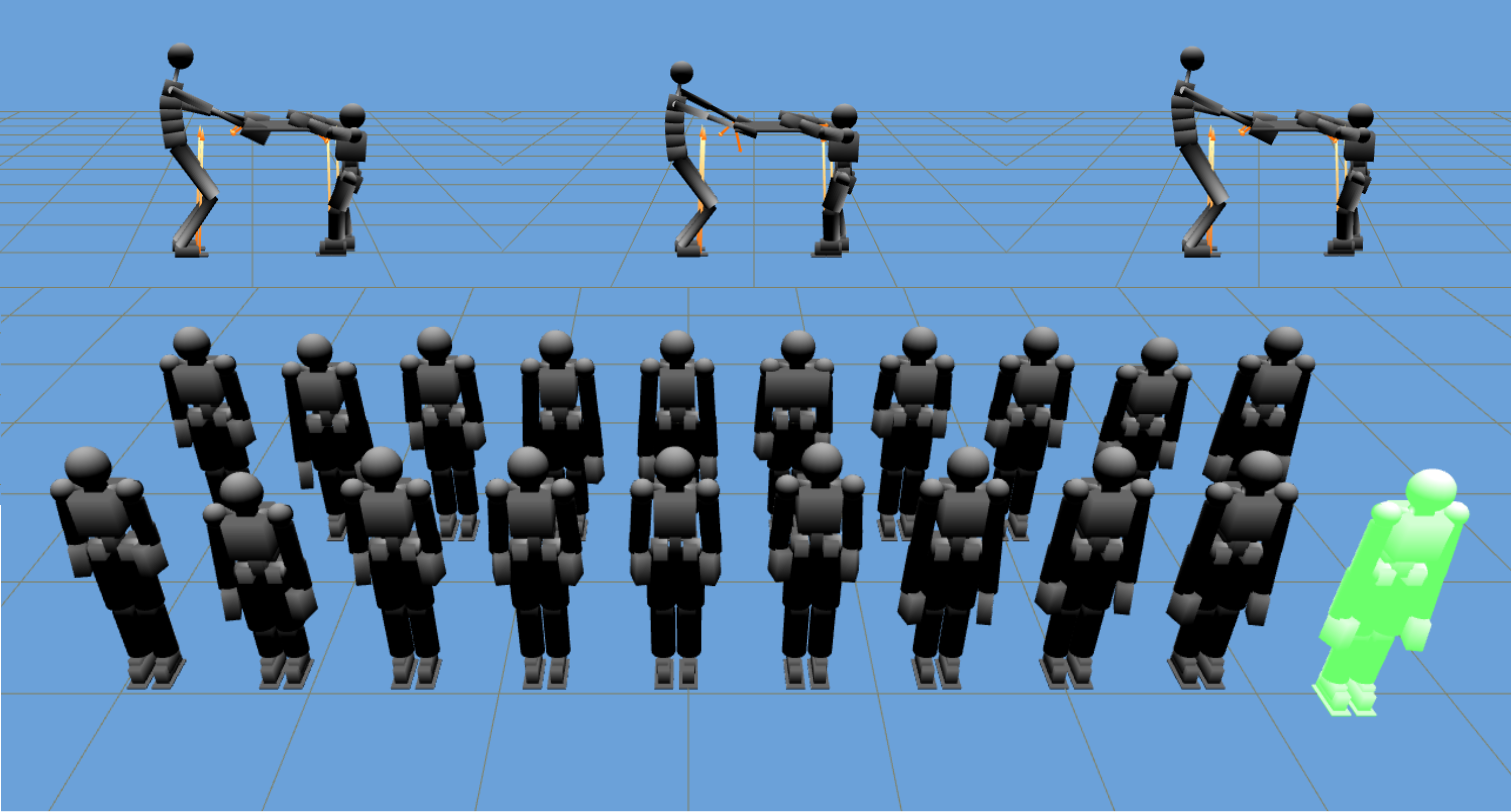}
    \caption{An example of a generation of the genetic algorithm. On the bottom part, a population is depicted, where, in green is identified the robot with the highest fitness.  The top part shows a candidate humanoid robot design collaborating with different humans in payload lifting tasks. Such tasks are used to evaluate the robot's fitness.}
    \label{fig:geneticOutput}
\end{figure}
The field of robotics has seen an increasing interest in the design and control of humanoid robots. These robots are expected to become part of our daily lives and collaborate with humans performing various tasks \cite{vaisi2022review}. The analysis of physical collaboration necessitates considering aspects such as human-centered ergonomics \cite{rasmussen2002design}, and safety and efficiency of the system \cite{alami2006}. The principles of \emph{ergonomics} have been widely integrated into the design of control architectures to achieve secure collaborations \cite{vysocky2016human, yokoyama2003cooperative}. However, the ergonomics of the agents have rarely been considered in the design of humanoid robot hardware. 
In this paper, we propose a novel methodology for designing humanoid robots by integrating ergonomic metrics through a bilevel optimization approach that combines genetic algorithms and nonlinear optimization.

In the field of robotics, co-design has been a critical area of research for optimizing the design of robots for specific tasks. Three primary methodologies have been applied, comprising \emph{classical optimization methods}, \emph{reinforcement learning}, and \emph{genetic algorithms}. Classical optimization methods have been widely used to optimize together hardware and control of robots considering both the kinematic and dynamic of the mechanism \cite{mombaur_longman_bock_schlöder_2005},\cite{allison2013multidisciplinary}. In addition, techniques such as Alternating Direction Method of Multipliers (ADMM) have been explored to expand the number of tasks considered in the optimization process \cite{bravo2022large}. Nevertheless, the complexity of humanoid robots and possible interactions with humans have not been taken into account. In \cite{sartore2022optimization}, we propose the use of nonlinear optimization techniques to design humanoid robots to maximize the ergonomics of the interaction with humans.  Regardless, a single human was considered and the methodology was sensitive to the problem's initial condition. In addition, only the links' characteristics were optimized w.r.t. the collaborative tasks. Reinforcement Learning techniques have shown promising results in the problem of optimizing robot design \cite{ha2019reinforcement},\cite{Bjelonic2023} considering the hardware as part of the policy and showing how the task directly affects the design of the robot. Yet, the designs considered were composed of a limited number of links, and humans were not considered in the loop. The use of Genetic Algorithms to achieve multi-objective optimization is recently gaining popularity \cite{bhatia2021evolution}. In \cite{9349280} behavior and mechanism are optimized together via the usage of genetic algorithm, meanwhile in \cite{fadini2022simulation}, bilevel optimization is used to design together control and hardware to minimize energy consumption and ensure control robustness. 
However, \cite{fadini2022simulation} consider platforms with limited degrees of freedom and do not account for possible interaction with a human and the related ergonomics requirements.

In this work, we introduce a novel method for optimizing robot design using genetic algorithms and nonlinear optimization, taking into account the complexity of humanoid robot design and multiple interactions with humans. Building on our previous work \cite{sartore2022optimization}, this paper extends the scope of the methodology in several ways:

\begin{itemize}
    \item we include motor characteristics in the robot kinematics and dynamics parameterization, augmenting the set of hardware parameters considered;
    \item we analyze the collaboration performed with different humans and different loads in the design optimization; \looseness=-1
    \item we use genetic algorithms in combination with nonlinear optimization to avoid being sensitive to the problem's initial conditions.\looseness=-1
\end{itemize}
The proposed methodology will be used to drive the design of the ergoCub2 humanoid robot. The ergoCub2 robot is a humanoid robot designed in the context of the ergoCub project\footnote{\url{https://ergocub.eu/}} and is an evolution of the ergoCub robot. Such a robot is  aimed at physically collaborating with humans in both industrial and healthcare environments.
The design obtained with the bilevel optimization has been compared to the design found in \cite{sartore2022optimization}, both in static interaction and while performing a single agent payload lifting task. In the static case, the proposed design decreases the robot energy expenditure of $43\%$, and improves human ergonomics for the most critical load heights. In the payload lifting task, the robot energy expenditure is decreased by $38\%$, showing overall an improvement in the task ergonomics. 


\section{Background}
\label{sec:background}
\subsection{Notation}
\label{sec:background:notation}
The notation used in this paper is the following: 
\begin{itemize}
    \item $\mathcal{I}$ indicates the inertial reference frame.
    \item $\mathbb{1}_{n} \in \mathbb{R}^{n \times n}$ denotes an identity matrix of size $n$.
    \item $S(.) :\mathbb{R}^{3} \mapsto SO(3)$ denotes the \textit{skew-symmetric} operator, hence given two vectors $v,u \in \mathbb{R}^{3}$, it is defined as $v \times u = S(v)u$.
    \item  $\prescript{\mathcal{A}}{}{p}_{\mathcal{B}} = \begin{bmatrix}
        \prescript{\mathcal{A}}{}{x}_{\mathcal{B}} & \prescript{\mathcal{A}}{}{y}_{\mathcal{B}} & \prescript{\mathcal{A}}{}{z}_{\mathcal{B}}
    \end{bmatrix}^T \in \mathbb{R}^3$ is the position of the origin of the frame $\mathcal{B}$ w.r.t. the frame $\mathcal{A}$.
    \item  $\prescript{\mathcal{A}}{}{R}_{\mathcal{B}} \in SO(3)$ is the rotation matrix of a frame $\mathcal{B}$ w.r.t. a frame $\mathcal{A}$.
    \item The operator $\left\lVert . \right\rVert_2$ indicates vector squared norm. Given $v \in \mathbb{R}^{n}$, it is defined as $\left\lVert v \right\rVert_2 = \sqrt{v_1^2+...+v_n^2}$. 
    \item  $\prescript{\mathcal{A}}{}{\phi}_{\mathcal{B}} \in \mathbb{R}^4$ with  $\left \rVert \prescript{\mathcal{A}}{}
    {\phi}_{\mathcal{B}}\right \rVert_2= 1$  is the quaternion representing the rotation of a frame $\mathcal{B}$ w.r.t. a frame $\mathcal{A}$, to convert the quaternion to the rotation matrix, the Rodrigues formula \cite{murray2017mathematical} is used.
    \item $\prescript{\mathcal{A}}{}{\omega}_{\mathcal{B}} \in \mathbb{R}^3$ is the angular velocity expressed in $\mathcal{A}$ of the frame $\mathcal{B}$ w.r.t. the frame $\mathcal{A}$, i.e. $S(\prescript{\mathcal{A}}{}{\omega}_{\mathcal{B}}) = \prescript{\mathcal{A}}{}{\dot{R}}_{\mathcal{B}} \prescript{\mathcal{A}}{}{R}_{\mathcal{B}}^\top$. 
\end{itemize}
  \subsection{Robot Modelling}
 \label{sec:background-robot-modelling}
 A humanoid robot is a mechanical system composed of $n+1$ rigid bodies, namely the \textit{links}, that are connected via $n$ \textit{joints}, each one endowed with one {degree of freedom} (DoF). The robot is represented as a {floating base} system, hence none of the links has an {a priori} constant \emph{position} and \emph{orientation}, w.r.t. the inertial reference frame $\mathcal{I}$. We refer to the \textit{base frame}, denoted by $\mathcal{B}$, as a frame attached to a specific link of the chain. The \textit{model configuration} is represented by both the \emph{pose}, i.e. position-and-orientation, of the \textit{base frame},  and the \textit{joint positions}, and it belongs to the Lie group $\mathbb{Q}=\mathbb{R}^{3}\times SO(3) \times \mathbb{R}^{n}$. 
 An element of the configuration space $q \in \mathbb{Q}$ is defined as the triplet $q = (\prescript{\mathcal{I}}{}{p}_{\mathcal{B}}, \prescript{\mathcal{I}}{}{R}_{\mathcal{B}}, s)$ where $\prescript{\mathcal{I}}{}{p}_{\mathcal{B}} \in \mathbb{R}^3$ and $\prescript{\mathcal{I}}{}{R}_{\mathcal{B}} \in SO(3)$ denote the position and the orientation of the \textit{base frame} respectively, and $s \in \mathbb{R}^n$ is the joints configuration.
 It is possible to compute the position and orientation of a specific frame $\mathcal{A}$, attached to the kinematic chain, from the \textit{model configuration} via the geometrical forward kinematic map $h_{\mathcal{A}}(\cdot):\mathbb{Q} \mapsto \mathbb{R}^3\times SO(3)$. 
 The \textit{model velocity}, belonging to the velocity space $\mathbb{V} = \mathbb{R}^{6+n}$, is represented  by linear and angular velocity of the \textit{base frame} jointly with the \textit{joint velocities}. An element of the configuration velocity space $\nu \in \mathbb{V}$ is hence defined as $\nu = (\prescript{\mathcal{I}}{}{\mathrm{v}}_{\mathcal{B}}, \dot{s})$ where $\prescript{\mathcal{I}}{}{\mathrm{v}}_{\mathcal{B}}=(\prescript{\mathcal{I}}{}{\dot{p}}_{\mathcal{B}}, \prescript{\mathcal{I}}{}{\omega}_{\mathcal{B}}) \in \mathbb{R}^6$ denotes the linear and angular velocity of the \textit{base frame}, and $\dot{s}$ denotes the joint velocities. 
 The linear and angular velocity of a frame $\mathcal{A}$ attached to the kinematic chain, are represented by $\prescript{\mathcal{I}}{}{\mathrm{v}}_{\mathcal{A}}=(\prescript{\mathcal{I}}{}{\dot{p}}_{\mathcal{A}}, \prescript{\mathcal{I}}{}{\omega}_{\mathcal{A}})$ and can be computed via the \textit{Jacobian} ${J}_{\mathcal{A}}=\begin{bmatrix}
    {J}_{\mathcal{B},{\mathcal{A}}} & {J}_{s,{\mathcal{A}}} \end{bmatrix}^T \in \mathbb{R}^{6 \times (n+6)}$ which maps the system velocity $\nu$ into the frame velocity $\prescript{\mathcal{I}}{}{\mathrm{v}}_{\mathcal{A}}$, i.e.  $\prescript{\mathcal{I}}{}{\mathrm{v}}_{\mathcal{A}}= J_\mathcal{A}\nu$, with ${J}_{\mathcal{B},{\mathcal{A}}} \in \mathbb{R}^{6 \times 6}$ the base jacobian and ${J}_{s,{\mathcal{A}}} \in \mathbb{R}^{6 \times n} $ the joints jacobian.
\looseness=-1
\subsection{Parametrized Coupled Human-Robot-Load Dynamic}
\label{sec:background-parmametrization-human-robot-load-dynamic}
In \cite{sartore2022optimization} a parametrization of the floating base dynamics w.r.t. links hardware parameters has been proposed. 
First, a single rigid body dynamic is parametrized w.r.t. its density function $\rho  = \rho(r) : \mathbb{R}^3 \mapsto \mathbb{R}^+$ that maps each 3D point of the body $r$ to the density of that point, 
 and its geometry, represented by a set of parameters $l \in \mathbb{R}^{n_l}$. Then such a parametrization is extended to the case of multi-rigid body dynamics, by defining the set of hardware parameters considered $\beta$ that contains the hardware parameters for each link of the kinematic chain. 
 By applying the Euler-Poincarè formalism \cite{Marsden2010} the system dynamic is obtained and it is formed by a set of differential equations:
\looseness=-1
\begin{IEEEeqnarray}{RLC}
\label{eq:dynamic-divided}
& M_{\mathcal{B}}(q, \beta)\prescript{\mathcal{I}}{}{\dot{\mathrm{v}}}_{\mathcal{B}} {+} M_{\mathcal{B}s}(q,\beta)\dot{s} {+} h_{\mathcal{B}}(q,\nu, \beta) {=} J^T_{\mathcal{B}}f, \IEEEyessubnumber \label{eq:dynamic_base} \\
& M_s(q,\beta) \ddot{s} {+} M^T_{\mathcal{B}s}(q,\beta)\prescript{\mathcal{I}}{}{\dot{\mathrm{v}}}_{\mathcal{B}} {+} h_s(q,\nu, \beta) {=} J^T_sf {+}\tau, 
\IEEEyessubnumber  \label{eq:dynamic_joints}\\
\nonumber
\end{IEEEeqnarray}
with $M_{\mathcal{B}} \in \mathbb{R}^{6 \times 6}$, $M_{s} \in \mathbb{R}^{n \times n}$ and $M_{\mathcal{B}s} \in \mathbb{R}^{6 \times n}$ the components of the mass matrix, the terms $h_{\mathcal{B}} \in \mathbb{R}^{6}$ and $h_{s} \in \mathbb{R}^{n}$ account for the Coriolis and gravity effects and they refer to, respectively, the base and the joints, ${\tau}  \in \mathbb{R}^{n}$ is a vector representing the robot's joint torques, $f \in \mathbb{R}^{6n_c}$ represents the wrenches acting on $n_c$ contact links of the robot, and $J_{\mathcal{B}} \in \mathbb{R}^{6 \times 6n_c}$  $J_{s} \in \mathbb{R}^{n \times 6n_c}$ are the  base and joint Jacobians of the contact frames.
Eq. \eqref{eq:dynamic-divided} can also be written in the compact form of Eq. \eqref{eq:constrained-dynamic-hardware-param}
\begin{equation}
\label{eq:constrained-dynamic-hardware-param}
M(q, \beta) \dot{\nu} + h(q,\nu,\beta) = B {\tau} + J_c^T(q,\beta) f,    
\end{equation}
where $B = \left[0,\mathbb{1}_n\right]^T$ is a selector matrix and with  
\begin{equation}
    M(q,\beta) =
\begin{bmatrix}
    M_\mathcal{B}(q,\beta) & M_{\mathcal{B}s}(q,\beta) \\
    M^T_{\mathcal{B}s}(q,\beta)  & M_s(q,\beta)
\end{bmatrix},
\nonumber
\end{equation}
\begin{equation}
\begin{split}
    h(q,\nu, \beta) =
\begin{bmatrix}
    h_\mathcal{B}\\
    h_s
\end{bmatrix} &, \text{  } J_c (q,\beta) =
\begin{bmatrix}
    J_\mathcal{B} \\
    J_s
\end{bmatrix}.
\end{split}
\end{equation}
Starting from Eq. \eqref{eq:constrained-dynamic-hardware-param} and the work done in \cite{tirupachuri2019}, the coupled human-robot-load dynamics parametrized w.r.t. the links' hardware parameters can be defined as: 
\begin{equation}
\begin{split}
\label{eq:multi-system-equations}
& \begin{bmatrix} M_1(q_1, \beta) & 0 & 0\\\ 0 & M_2(q_2) & 0 \\ 0 & 0 & M_3(q_3)\end{bmatrix} \begin{bmatrix} \dot{\nu}_1 \\ \dot{\nu}_2 \\ \dot{\nu}_3 \end{bmatrix} + \begin{bmatrix} h_1(q_1,\nu_1,\beta) \\ h_2(q_2,\nu_2) \\h_3(q_3, \nu_3) \end{bmatrix} =\\ & \begin{bmatrix} B_1 & 0 \\\ 0 & B_2 \\\ 0& 0 \end{bmatrix}  \begin{bmatrix} \tau_1 \\\ \tau_2 \end{bmatrix} + \mathbf{Q}(q_1,q_2,q_3, \beta)^T \mathbf{f}, \\
& {\mathbf{\dot{Q}}}(q_1,q_2,q_3,\beta) \begin{bmatrix} \nu_1 \\ \nu_2 \\ \nu_3 \end{bmatrix} +\mathbf{Q}(q_1,q_2,q_3,\beta) \begin{bmatrix} \dot{\nu}_1 \\ \dot{\nu}_2 \\ \dot{\nu}_3 \end{bmatrix} = 0,
\end{split}
\end{equation}
where subscript $1$ refers to the human quantity, subscript $2$ to the robot one, and subscript $3$ to the load. In Eq. \eqref{eq:multi-system-equations}, the \textbf{bold} symbols identify the composite matrices, that consider quantities related to the load and both agents, i.e. the human and the robot. $\mathbf{Q}$ is a coupling matrix considering both the constraints of the contacts with the environment ($J^{e}\nu=0$) and those of the agents' load contact points, hence ($J_1^{i}\nu_1=J_3^{i}\nu_3$, $J_2^{i}\nu_1=J_3^{i}\nu_3$). The ordering in the matrix $\mathbf{Q}$, is reflected in  $\mathbf{f}$ which contains all the interaction wrenches (exchanged with the environment and between the agents and the payload) taking into account the action-reaction property for internal forces ($f^{i}_1=-f^{i}_3$, $f^{i}_2=-f^{i}_3$).
\subsection{Genetic Algorithms}
\label{sec:background:geneticAlgorithm}
\emph{Genetic algorithms} 
are a class of optimization algorithms inspired by the natural evolution process and genetics \cite{goldberg1988holland}. The basic idea behind genetic algorithms is to model the process of evolution by selecting the fittest individuals in a population and crossbreeding them to create a new generation of individuals with improved fitness. The genetic algorithm starts from an initial population of chromosomes, composed of different genes representing the chromosome's characteristics.  These chromosomes are evaluated based on a \emph{fitness function} that measures their suitability as a solution to the problem. The aim of the algorithm is to find the chromosome that maximizes the fitness function by evolving the population. The population evolution is performed via three principal methods, namely \emph{reproduction}, \emph{crossover}, and \emph{mutation}. Reproduction is aimed at selecting chromosomes, the so-called parents, from the population, based on their fitness value. Different reproduction method exists in the literature, such as the roulette wheel selection \cite{lipowski2012roulette} and the k-tournament \cite{goldberg1991comparative}. The crossover method is then used to generate new chromosomes, by mixing the parents' genes. The mutation is used to randomly change the chromosome genes. This introduces diversity into the population and helps to prevent the algorithm from getting stuck in local optima. 
\section{Bilevel Optimization of Humanoid Robots with Ergonomic Metrics }
\label{sec:contribution}
In this Section, we describe the methodology used to optimize the design of humanoid robots based on ergonomic metrics that consider interactions with multiple humans. First, in Sec. \ref{sec:contribution-motor-parametrization}, we introduce a parametrization of the robot's kinematics and dynamics w.r.t. hardware parameters, extending the parametrization presented in Sec. \ref{sec:background-parmametrization-human-robot-load-dynamic} by including motor characteristics. Then, in Sec. \ref{sec:contribution:genetic_algorithm}, we present a bilevel optimization approach that leverages the proposed parametrization to identify optimal robot designs.
\subsection{Multi Rigid Body Parametrization w.r.t. Motor Characteristics\looseness=-1  }
\label{sec:contribution-motor-parametrization}
In this Section, we expand the hardware parametrization of the robot dynamics and kinematics presented in Sec. \ref{sec:background-parmametrization-human-robot-load-dynamic} by considering the motor specifications for each actuated joint.

We identify with $\mathcal{N}$ the set of motors taken into account, to each element $\alpha_i$ of $\mathcal{N}$, different motor characteristics are associated, namely: an upper torque limit $\tau_{i}^+(\alpha_i) \in \mathbb{R}$, a lower torque limit $\tau_{i}^-(\alpha_i) \in \mathbb{R}$, a motor inertia $I_{i,m}(\alpha_i) \in \mathbb{R}$, a gear-box ratio $\Gamma_{i}(\alpha_i) \in \mathbb{R}$, and a viscous friction coefficient $K_{i,v}(\alpha_i) \in \mathbb{R}$. 
To each of the motors of the kinematic chain, an element $\alpha_i \in \mathcal{N}$ has been associated and collected in  $\alpha \in \mathcal{N}^n$. Consequently, the following matrices are defined considering the whole kinematic chain: 
 $\Gamma(\alpha) \in \mathbb{R}^{n\times n}$, i.e., the matrix accounting for the gear-box ratio and the coupling between the input and output rotations of the coupling mechanism. $I_m(\alpha) \in \mathbb{R}^{n \times n}$ is the motors inertia matrix. $\tau^-(\alpha),\tau^+(\alpha)  \in \mathbb{R}^n$ are the lower and upper motors torque limit, respectively.  $K_v(\alpha) \in \mathbb{R}^{n \times n}$ is the matrix collecting the viscous friction coefficient. 

By assuming that each joint is rigidly connected to the related motors by the transmission element, the relationship between the position of the joints $s \in \mathbb{R}^n$ and the position of the motors $\theta \in \mathbb{R}^n$  can be written as a function of the motor type vector $\alpha$ as $s = \Gamma(\alpha) \theta$.
Then, analogously to the work done in \cite{nava2018exploiting}, we assume that the friction is modeled as viscous friction only and that the angular motor kinetic energy is due to its spinning only. In addition, the center of mass of each motor is assumed to be along the motor axis of rotation. Starting from those assumptions, we can write the motor dynamic as a function of the motor state, $\theta, \dot{\theta}, \ddot{\theta}$ and the motor type vector $\alpha$ as per the following equation:
\begin{equation}
 \label{eq:motor-dynamics}
      I_m(\alpha)\ddot{\theta} + K_{v}(\alpha)\dot{\theta} = \tau_m - \Gamma(\alpha)\tau.
 \end{equation}
Where $\tau_m \in \mathbb{R}^n$ are the motors' torque. By unifying the relationship between motor and joint position with Eq. \eqref{eq:motor-dynamics}, and substituting $\ddot{\theta}$ with $\ddot{s}$ we can write Eq. \eqref{eq:motor-joints-dynamics}.
\begin{equation}
 \label{eq:motor-joints-dynamics}
      I_m(\alpha)\Gamma(\alpha)^{-1}\ddot{s} = \tau_m - \Gamma(\alpha)^{-1}\tau - K_{v}\Gamma(\alpha)^{-1}\dot{s},
 \end{equation}
then by multiplying Eq. \eqref{eq:motor-joints-dynamics} by $\Gamma(\alpha)^{-T}$ and summing Eq. \eqref{eq:dynamic_joints} with Eq. \eqref{eq:motor-joints-dynamics} we get: 
\begin{equation}
\begin{split}
\label{eq:dynamics_joints_motor_coupled}
& \bar{M}_s(q,\beta,\alpha) \ddot{s} {+} {M}^T_{\mathcal{B}s}(q,\beta,\alpha)\prescript{\mathcal{I}}{}{\dot{\mathrm{v}}}_{\mathcal{B}}{+}h_s(q,\nu,\beta) {=}\\ 
& J_s(q,\beta)^{T}f {-} \bar{K}_v(\alpha)\dot{s} {+} \Gamma(\alpha)^{-T} \tau_m,
\end{split}
\yesnumber
\end{equation}
 with $\bar{M}_s(q,\beta,\alpha) = M_s(q,\beta) + \Gamma(\alpha)^{-T}I_m(\alpha)\Gamma(\alpha)^{-1},$ and $\bar{K}_v(\alpha) = \Gamma(\alpha)^{-T}K_v(\alpha)\Gamma(\alpha)^{-1}.$
By defining a new set of hardware parameters $\pi = \begin{bmatrix}
   \beta & \alpha 
\end{bmatrix}^T$ and unifying Eq. \eqref{eq:dynamic_base} with  Eq. \eqref{eq:dynamics_joints_motor_coupled}, we can write the floating base dynamic together with the motor dynamics, parametrized w.r.t. links' and motors' parameters as $\bar{M}(q, \pi) \dot{\nu} {+} h(q,\nu,\pi) {=} \bar{B}(\pi) {\tau_m} {+} J_c^T(q,\pi) f {-}\bar{K_v}(\pi)\nu$
with: 
\begin{equation}
    \label{eq:mass_matrix_bar}
    \bar{M}(q,\pi) = \begin{bmatrix}
    M_\mathcal{B}(q,\pi) & {M}_{\mathcal{B}s}(q,\pi)\\
    {M}^T_{\mathcal{B}s}(q,\pi) & \bar{M}_s(q,\pi)    
    \end{bmatrix},
\nonumber
\end{equation}
\begin{equation}
\label{eq:b_bar}
\begin{split}
    \bar{B}(\pi) = \begin{bmatrix}
    0 \\
    \Gamma(\pi)^{-T}
    \end{bmatrix} &, \text{  } \bar{K_v}(\pi) = \begin{bmatrix}
    0 \\
    K_v(\pi)
    \end{bmatrix}.
\end{split}   
\nonumber
\end{equation}
Then, the coupled human-robot-load dynamic of Eq. \eqref{eq:multi-system-equations} can be re-written to consider also the motor dynamics as: 
\begin{equation}
\begin{split}
\label{eq:multi-system-equations-motor-dynamics}
& \begin{bmatrix} \bar{M}_1(q_1, \pi) & 0 & 0\\\ 0 & M_2(q_2) & 0 \\ 0 & 0 & M_3(q_3)\end{bmatrix} \begin{bmatrix} \dot{\nu}_1 \\ \dot{\nu}_2 \\ \dot{\nu}_3 \end{bmatrix} {+}\begin{bmatrix} h_1(q_1,\nu_1,\pi) \\ h_2(q_2,\nu_2) \\h_3(q_3, \nu_3) \end{bmatrix} {=}\\ & \begin{bmatrix} \bar{B}(\pi)_1 & 0 \\\ 0 & B_2 \\\ 0& 0 \end{bmatrix}  \begin{bmatrix} \tau_{m,1} \\\ \tau_2 \end{bmatrix} {+} \mathbf{Q}(q_1,q_2,q_3, \pi)^T \mathbf{f} {-} \begin{bmatrix}
    0 \\ \bar{K_v} \\ 0
\end{bmatrix}\begin{bmatrix} \nu_1 \\ \nu_2 \\ \nu_3 \end{bmatrix} , \\
& {\mathbf{\dot{Q}}}(q_1,q_2,q_3,\pi) \begin{bmatrix} \nu_1 \\ \nu_2 \\ \nu_3 \end{bmatrix} {+}\mathbf{Q}(q_1,q_2,q_3,\pi) \begin{bmatrix} \dot{\nu}_1 \\ \dot{\nu}_2 \\ \dot{\nu}_3 \end{bmatrix} {=} 0,
\end{split}
\end{equation}
thus obtaining the human-load-robot coupled dynamic parametrized w.r.t. the motors and links hardware parameters. 
For the sake of simplicity, we can rewrite Eq. \eqref{eq:multi-system-equations-motor-dynamics} in its compact form, as: 
\begin{equation}
\label{eq:multi-system-equations-compact-hardware}
\begin{split}
& \mathbf{\bar{M}} (\boldsymbol{q}, \pi){\boldsymbol{\dot{\nu}}} {+} \mathbf{h}(\boldsymbol{q}, \boldsymbol{\nu},\pi) {=} \mathbf{\bar{B}}(\pi) \bar{\boldsymbol{\tau}} {+} \mathbf{Q}^T(\boldsymbol{q}, \pi) \mathbf{f} {+} \mathbf{\bar{K}_v}(\pi) \boldsymbol{\nu}, \\
& {\mathbf{\dot{Q}}}(\boldsymbol{q}, \pi) \boldsymbol{\nu} +\mathbf{Q}(\boldsymbol{q}, \pi) \boldsymbol{\dot{\nu}} = 0.
\end{split}
\end{equation}
It is worth noticing as, in this formulation,  $\bar{\boldsymbol{\tau}} = \begin{bmatrix} \tau_{m,1} & \tau_2 \end{bmatrix}^T$, thus for the human the joint torques are considered, meanwhile the motor torques are considered for the robot.
\subsection{Bilevel Optimization Framework for Hardware Parameters Optimization}
\label{sec:contribution:genetic_algorithm}
In this Section, we describe the bilevel optimization framework for identifying the optimal set of robot hardware parameters $\pi$, defined as per Sec. \ref{sec:contribution-motor-parametrization}, to improve the postural ergonomics of payload lifting tasks performed with various users and loads. The framework proposed utilizes a genetic algorithm to generate a population of robot designs. Nonlinear optimization problems are then employed, based on the parametrization presented in Sec. \ref{sec:contribution-motor-parametrization}, to evaluate the ergonomic metrics of the interaction between the robot designs and different users and loads. The fitness associated with each robot design is computed using the results of the optimization problems and it is used to update and evaluate the genetic algorithm population.


\begin{figure}[!t]
\centering
    \includegraphics[trim=0.0cm 0.0cm 0.0cm 0.0cm, clip=true, width=0.95\columnwidth]{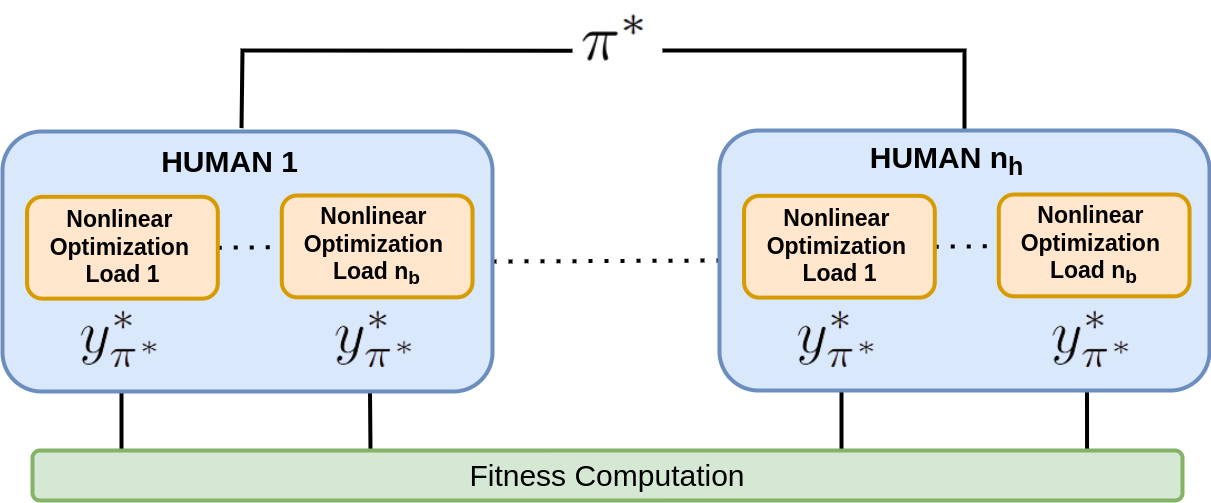}
    \caption{The fitness function computation. Given a chromosome, represented by a set of hardware parameters $\pi^*$, different nonlinear optimization problems are solved, one per human-load combination considered. The outputs of such optimizations are used to compute the fitness function associated with the given chromosome.
    }
    \label{fig:diagram}
\end{figure}
Specifically, the genetic algorithm generates a population of $n_g$ chromosomes, as shown in Fig. \ref{fig:geneticOutput}. Each chromosome identifies a different set of hardware parameters $\pi$, composed by links geometry, densities and joint motor types. The chromosomes generated are such that the hardware parameters are feasible, hence $l \in \mathbb{R}^{n_l+}$, $\rho \in \mathcal{P}$ where $\mathcal{P}$ includes a set of feasible densities associated to different materials, and $\alpha \in \mathcal{N}^n$ the set of motors type defined in Sec. \ref{sec:contribution-motor-parametrization}. 
The fitness function associated with each chromosome considers the ergonomy of different collaborations.
We evaluate the ergonomy of the interactions using the metric of \emph{postural ergonomy} \cite{sartore2022optimization}. Energy expenditure is computed for both agents based on internal torques, following a similar approach in \cite{Rapetti2021}. We specifically chose this metric due to its representative nature in capturing agents' energy expenditure.
To determine the fitness and compute internal torques during collaborations, we formulate various optimization problems, as shown in Fig. \ref{fig:diagram}. Our approach considers both human and robot energy expenditure, striking a balance between optimizing the ergonomic experience for the human collaborator and minimizing torque requirements for the robot.
Each optimization problem considers $n_k$ load height and it computes the $n_k$ robot, human, and load configurations that minimize the internal torques and allow holding the load at the given heights, in static conditions. 
More into detail, given a robot design identified by the chromosome $\pi^*$, for each human-load combination, we define the following \textbf{optimization problem}.  
\subsubsection{Search Variable}
The search variable of the optimization problem is equal to $y = \begin{bmatrix}
 &\boldsymbol{\tilde{q}} &\boldsymbol{\tilde{\nu}} & \boldsymbol{\tilde{\dot{\nu}}} &\bar{\boldsymbol{\tilde{\tau}}} & \boldsymbol{\tilde{f}} & 
\end{bmatrix}^T$ where $\tilde{\cdot}$ represents a set of variables, one per target object height considered, $\boldsymbol{q}$, $\boldsymbol{\nu}$, $\boldsymbol{\dot{\nu}}$ are, respectively, the human-robot-load configuration, velocities and accelerations. $\bar{\boldsymbol{\tau}}$ collects the robot motor torques and the human joint torques and $\boldsymbol{f}$ are the interaction wrenches exchanged with the environment and between the agents and the payload. 
\subsubsection{Dynamics Constraints}
The first constraint introduced, consists of the coupled human-robot-load dynamics of Eq. \eqref{eq:multi-system-equations-compact-hardware} computed with $\pi = \pi^*$, hence: 
\begin{equation}
\label{eq:dynamic-constraints}
\begin{split}
& \mathbf{\bar{M}} (\boldsymbol{q}_k, \pi^*){\boldsymbol{\dot{\nu_k}}} {+} \mathbf{h}(\boldsymbol{q_k}, \boldsymbol{\nu_k},\pi^*) =\mathbf{\bar{B}}(\pi^*) \bar{\boldsymbol{\tau_k}} {+} \\
&{+}\mathbf{Q}^T(\boldsymbol{q_k}, \pi^*) \mathbf{f_k} {+} \mathbf{\bar{K}_v}(\pi^*) \boldsymbol{\nu_k},  \text{ } \forall k \in \boldsymbol{K}\\
& {\mathbf{\dot{Q}}}(\boldsymbol{q}_k, \pi^*) \boldsymbol{\nu_k}{+}\mathbf{Q}(\boldsymbol{q_k}, \pi^*) \boldsymbol{\dot{\nu_k}} = 0  \text{ } \forall k \in \boldsymbol{K} 
\end{split}
\end{equation}
where $\boldsymbol{K} = [1,n_k]$.
Since static configurations are considered, the following constraints are introduced, which force the system velocities and acceleration to be null, for each target load height considered: 
\begin{equation}
\label{eq:null-vel-and-acc-constraints}
\begin{split}
& \boldsymbol{\dot{\nu}}_k =0 \text{ } \forall k \in \boldsymbol{K}\\
& \boldsymbol{\nu_k}= 0 \text{ } \forall k \in \boldsymbol{K}
\end{split}
\end{equation}
To ensure that the interaction is feasible from a dynamic point of view, the interaction wrenches should be inside the friction cones, resulting in the following constraint:  \looseness=-1 
\begin{equation}
\label{eq:friction-cones}
\begin{split}
& C\boldsymbol{f}_k<b  \text{ } \forall k \in \boldsymbol{K}
\end{split}
\end{equation}
In addition, the robot motors torque should be inside the limits identified by the set of hardware parameters $\pi^*$, resulting in the following constraint:
\begin{equation}
\label{eq:torques-limits-constraints}
\begin{split}
& \tau^-(\pi^*)<\tau_{m,1,k}<\tau^+(\pi^*)  \text{ } \forall k \in \boldsymbol{K}
\end{split}
\end{equation}
\subsubsection{Kinematic Constraints}
To ensure the feasibility of the interaction,  the kinematic constraint of Eq. \eqref{eq:kinematic-constraints} should be satisfied, where, for the sake of simplicity, the dependence on the search variable $y$ and the set of hardware parameters $\pi^*$ has been ommitted.  
\begin{IEEEeqnarray}{RLC}
\IEEEyesnumber
\label{eq:kinematic-constraints}
& \prescript{\mathcal{I}}{}{p}_{{c,i,k}} = \prescript{\mathcal{I}}{}{p}_{{c,3,k}} \text{ } \forall i \in [1,2], \text{ } \forall c \in \mathcal{F}_c, \text{ } \forall k \in \boldsymbol{K} \IEEEyessubnumber \label{eq:pose-constraint} \\
& \prescript{\mathcal{I}}{}{R}_{{c,k}} = \prescript{\mathcal{I}}{}{R}^*_{{c,k}} \text{ } \forall c \in \mathcal{F}_R, \text{ } \forall k \in \boldsymbol{K} \IEEEyessubnumber \label{eq:rotation-constraint} \\
& \prescript{\mathcal{I}}{}{z}_{{c,k}} = \prescript{\mathcal{I}}{}{z}^*_{{c,k}} \text{ } \forall c \in \mathcal{F}_R, \text{ } \forall k \in \boldsymbol{K} \IEEEyessubnumber \label{eq:height-constraint}\\
& \prescript{\mathcal{I}}{}{x}_{{3,k}} = \prescript{\mathcal{I}}{}{x}_{{3,k+1}} \text{ }, \text{ } \forall k \in \boldsymbol{K'} \IEEEyessubnumber \label{eq:subsequent-x-constraint}\\
& \prescript{\mathcal{I}}{}{y}_{{3,k}} = \prescript{\mathcal{I}}{}{y}_{{3,k+1}} \text{ }, \text{ } \forall k \in \boldsymbol{K'} \IEEEyessubnumber \label{eq:subsequent-y-constraint}\\
& \prescript{\mathcal{I}}{}{p}_{{c,k}} = \prescript{\mathcal{I}}{}{p}_{{c,k+1}} \text{ }  \forall c \in \mathcal{F}_p, \text{ } \forall k \in \boldsymbol{K'} \IEEEyessubnumber \label{eq:subsequent-pose-constraint} 
\end{IEEEeqnarray}
Where  $\boldsymbol{K'} = [1,n_k-1]$, $\mathcal{F}_c$ of Eq. \eqref{eq:pose-constraint} contains the hand frames for both agents, $\mathcal{F}_R$ of Eq. \eqref{eq:rotation-constraint} and Eq. \eqref{eq:height-constraint} contains the box frame and the feet frames for both agents, and $\mathcal{F}_p$ of Eq. \eqref{eq:subsequent-pose-constraint} contains the feet frames for both agents.
More specifically, Eq. \eqref{eq:pose-constraint} forces the hands of both agents to be at the correct position on the load, Eq. \eqref{eq:rotation-constraint} constrains the load orientation and the agents feet orientation to a reference value, defined such that they are parallel to the ground, Eq. \eqref{eq:height-constraint} forces the box to be at the target height considered and the feet to be on the ground, Eq. \eqref{eq:subsequent-x-constraint} and Eq. \eqref{eq:subsequent-y-constraint} forces the load not to move along the x and y directions in between the different load heights considered and Eq. \eqref{eq:subsequent-pose-constraint} forces both agents' feet to remain stationary throughout the different load heights considered. 
\subsubsection{Intrinsic Constraints}
A set of constraints is introduced to ensure that the variable found is consistent with its physical meaning.\looseness=-1
\begin{IEEEeqnarray}{RLC}
\IEEEyesnumber
\label{eq:intrinsic-constraint}
&  \left\rVert \prescript{\mathcal{I}}{} \phi_{c,k}  \right\rVert_2^2 = 1 \text{ } \forall c \in \mathcal{F}_\phi, \text{ } \forall k \in \boldsymbol{K}\IEEEyessubnumber \label{eq:quaternion-norm} \\
& A_{i}s_{i,k} = 0\text{ } \forall i \in [1,2]  \text{ } \forall k \in \boldsymbol{K} \IEEEyessubnumber \label{eq:symmetry-joints} \\
& s^-_i< s_{i,k}< s^+_i \text{ } \forall i \in [1,2] \text{ } \forall k \in \boldsymbol{K} \IEEEyessubnumber
\label{eq:joints-limits}
\yesnumber
\end{IEEEeqnarray}
Where $\mathcal{F}_\phi$ in Eq. \eqref{eq:quaternion-norm} contains the box frame and both agents' base frame. The constraint of Eq. \eqref{eq:quaternion-norm}, forces the quaternion of the frames $\in \mathcal{F}_{\phi}$ to be of unitary norm; anyhow, since the solver might perform intermediate unfeasible iterations, the quaternions are normalized every time the frames are evaluated.  
In Eq. \eqref{eq:symmetry-joints}, $A_i \in \mathbb{R}^{n \times n}$ is defined such that both the  human's and robot's joint configurations are symmetrical .  Eq. \eqref{eq:joints-limits} ensures that the human and robot joint configurations found are within the limits.  
\subsubsection{Optimization Problem}
The following optimization problem is eventually defined: 
\begin{IEEEeqnarray}{RCL}
\label{eq:optimization}
\IEEEyesnumber
 y^* & = & \underset{y}{\text{argmin}}\left( \sum_{k=1}^{n_k}\left( W_1t_{1,k}{+}W_2t_{2,k}\right){+} \sum_{k=1}^{n_k-1}W_3t_{3,k}\right) \nonumber  \\
 &\text{s.t. } & \nonumber\\
    && \text{Coupled Human-Load-Robot Dynamics of Eq. }\eqref{eq:dynamic-constraints} \nonumber\\
    && \text{Static Conditions of Eq. }\eqref{eq:null-vel-and-acc-constraints}\nonumber \\
    && \text{Wrenches Constraints of Eq. } \eqref{eq:friction-cones} \nonumber \\
    && \text{Torque Limits of Eq. } \eqref{eq:torques-limits-constraints} \nonumber \\
    && \text{Kinematic Constraints of Eq. }\eqref{eq:kinematic-constraints} \nonumber\\
    && \text{Intrinsic Constraints of Eq. }\eqref{eq:intrinsic-constraint} \nonumber\\
\yesnumber
\end{IEEEeqnarray}
with
\begin{itemize}
    \item $t_{1,k} = \left \rVert w_1^T\bar{\boldsymbol{\tau}}_k \right\rVert_2^2$ a task minimizing the human joints and the robot motors torque. Since the human joint torque and robot motor joint torque have different scales, a weight $w_1$ is introduced to appropriately consider both contributions. \looseness=-1
    \item $t_{2,k} = \left\rVert s_{i,k}-s^*_{i,k}\right\rVert_2^2 \text{  } \forall i \in [1,2] $ this task regularizes the joint configurations of both agents to avoid singular positions, preventing the optimizer from selecting low-torque singular configurations. It also helps avoid making design choices that overly focus on singular configurations, which can lead to suboptimal performance in practical scenarios.
    \item $t_{3,k}= \left\rVert\boldsymbol{q}_{i,k}-\boldsymbol{q}_{i,k+1}\right\rVert_2^2 \text{ } \forall i \in [1,2]$ a task minimizing the difference between the configuration associated to different loads height, for both agents.  
\end{itemize}
The interaction between different components of the cost function should be considered. Properly selecting weights (e.g., $W_1, W_2, W_3$) ensures that minimizing energy expenditure remains the primary objective, while the other tasks act as regularization terms. By considering $n_h$ humans and $n_o$ different loads, we will solve $n_h \times n_o$ different optimization problems, each one considering $n_k$ load heights, obtaining  $n_h \times n_o \times n_k$ human and robot torques associated to each chromosome. We can then compute the chromosome fitness as $f_{\pi^*} = W_f\frac{1}{\left \rVert\bar{\boldsymbol{\tau}}_w\right \rVert_2}$,
where $\bar{\boldsymbol{\tau}}_w$ is the torque with the biggest norm between the  $n_h \times n_o \times n_k$ torques vector computed with the nonlinear optimizations. In this way, we consider the worst case, from an  energy expenditure  point of view, in between the ones considered, thus avoiding that a robot performing exceptionally well in one collaboration but badly in another one  will be chosen over a robot performing in average well with all the humans. $W_f \in \mathbb{R}$ is a weight aimed at modulating the fitness.
In addition, since the genetic algorithm aims at maximizing chromosome fitness, the reciprocal of the worst torque norm is used, since we want to minimize the internal energy expenditure.
In the fitness both human and robot metrics appear since both human joint and robot motor torques are included in the vector $\bar{\boldsymbol{\tau}}_w$.
\begin{table}[!t]
\begin{center}
\caption{The set of motors taken into account, with the associated characteristics. The torque limits in this table, refer to joint torque, hence after the transmission.}
\label{tab:motors}
\begin{tabular}{|c|c|c|c|c|c|}
\hline
Motor ID &$\frac{1}{\Gamma_i}$& $I_{m,i} [\kilogram \meter^2]$&$\tau_{i}^-[N\meter]$ & $\tau_{i}^+[N\meter]$ \\ \hline
S & $100$ & $10^{-4}$ & $-37$ & $37$ \\ \hline
M & $160$ & $10^{-3}$ & $-92$ & $92$  \\ \hline
L & $160$ & $10^{-3}$ & $-123$ & $123$  \\ \hline
\end{tabular}
\end{center}
\end{table}
\section{Results}
\label{sec:validation}
This Section is divided into three parts. In the first part, Sec. \ref{sec:results:validation-settings}, we describe the implementation and validation of the proposed methodology, including details about the behavior of the genetic algorithm. Then we compare the robot design obtained using the proposed approach to a design obtained using nonlinear optimization.
In Sec. \ref{sec:validation:staticCase}, we compare the robots in a static task. Thus we evaluate the ergonomy of a collaboration with a human in stationary conditions.
Meanwhile, in Sec. \ref{sec:validation:simulation}, we compare the designs in a dynamic task, by computing the robots energy expenditure when performing a payload lifting task.
\begin{figure}[!t]
\centering
 \includegraphics[trim=3cm 1.0cm 3cm 2cm, clip=true,width=\columnwidth]{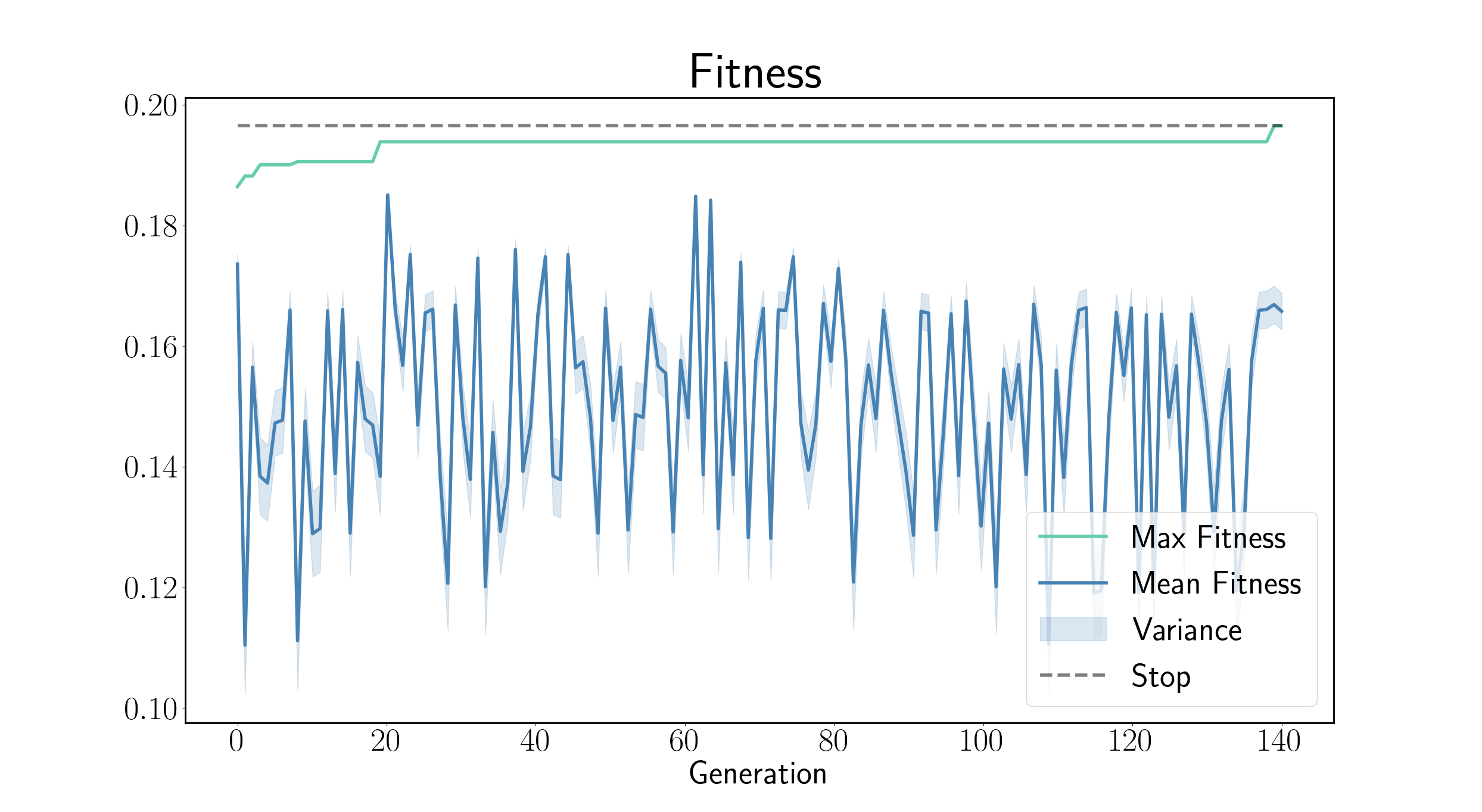}
\caption{The mean, the variance, and the maximum of the population fitness w.r.t. the generation. The dashed line identifies the fitness threshold for stopping the genetic algorithm.\looseness=-1}
\label{fig:fitness}
\end{figure}

\subsection{Validation}
\label{sec:results:validation-settings}
The dynamics parametrization and hardware optimization proposed in Sec. \ref{sec:contribution}, have been validated starting from the iCub humanoid robot \cite{metta2008icub}. Analogously to the work done in \cite{sartore2022optimization}, the iCub humanoid robot has been modeled with basic shapes, i.e. \textit{spheres}, \textit{cylinders}, and \textit{boxes}, to introduce the link parametrization of Sec. \ref{sec:background-parmametrization-human-robot-load-dynamic}. Each link's inertia characteristics have been parametrized w.r.t. $[l_m,\rho]$ where $l_m \in \mathbb{R}^+$ is a length multiplier that scales the link's geometry along the growing direction of the kinematic chain. $\rho $ is the link's density assumed to be constant within the link's volume. Each motor of the robot is associated with a motor type parameter, represented by $\alpha_i \in \mathcal{N}$. The set of motors $\mathcal{N}$ is depicted in  Table \ref{tab:motors}. These custom-built motors were specifically chosen and used for the design of the ergoCub2 robot to meet its design requirements and functional specifications. The optimization takes into account static configurations, therefore the viscous friction and velocity limits have not been considered since the velocities are forced to be zero. 
A total of $n_l=8$ length multipliers and $n_\rho=8$ densities have been considered and they are distributed as follows: $4$ in the torso, $2$ in the arms (one for the upper and one for the lower arms links) and $2$ for the legs, (one for the upper and one for the lower leg links). The symmetry in the robot shape is ensured by defining one set of links' characteristics for the arms and one for the legs. The total number of motors considered is  $n_m = 13$ distributed as follows: $3$ in the torso, $6$ in the legs actuating the hip, the ankle pitch and roll and one actuating the knee, and $4$ for the arms actuating the shoulder and the elbow. Also in this case symmetry is ensured by defining one set of motors for the arms and one for the legs. The total hardware parameters vector is composed as  $\pi = [l_{m1}, ..., l_{m_{n_l}}, \rho_1, ..., \rho_{n_\rho}, \alpha_1, ..., \alpha_{n_m}]^T$. \\
Three different humans have been considered, with height $1.68,1.78,1.82\meter$. These individuals were selected based on samples collected from real subjects, considering various ages and genders.
The humans have been modeled as a 48DoF multi-body system, as per \cite{latella2019simultaneous}. Such a model allows to compute human joint torques with accuracy, despite not considering the complexity of the musculoskeletal system. Thus improving the computational efficiency in solving the optimization problem while maintaining a realistic model of the human state \cite{latella2019simultaneous}.
Three different loads have been taken into account, with weights $5,10,15\kilogram$. The loads have all been modeled as a box of dimensions $0.5 \text{ }\meter \times 0.5 \text{ }\meter \times 0.025 \text{ }\meter$. 
Three different load heights have been considered, namely $0.8\meter,1.0\meter,$ and $1.2 \meter$.\\
The genetic algorithm has been implemented using \texttt{pyGAD}\cite{gad2021pygad}. Different genetic algorithms have been tested by varying the parameters of the evolutionary process. For the sake of brevity, only the settings that lead to the best outcome are shown. The genetic algorithm considers a population of $n_g=20$ robots, \emph{k-tournament} \cite{goldberg1991comparative} selection has been used for the offspring generation, with $k=3$ while random mutation has been used with a mutation gene percentage of 10\%. The algorithm was stopped once the fitness of the fittest chromosome improved by $5\%$. In our experience, this proved to be enough to have a meaningful improvement w.r.t. the starting point. 
The optimization problem of Sec. 
\ref{sec:contribution:genetic_algorithm}, used to compute the fitness, has been implemented using ADAM  \cite{l2022whole} with CasADi \cite{Andersson2019}, and \texttt{Ipopt}\cite{wachter2006implementation} with \texttt{MA27} \cite{HSL}  has been used as a solver. \\
The fitness function of the genetic algorithm population is shown in Fig. \ref{fig:fitness}. A total of $2885$ different robot models have been analyzed. As can be noticed, the algorithm converges towards a sub-optimal solution around generation $20$, then at generation $140$ a new optimal solution is found that meets the requirements of the stopping criteria.  A warm start strategy was employed in the genetic algorithm to address the large search space. It included outputs from the nonlinear optimization approach and specific characteristics of the iCub3 robot. The fitness function trend suggests that a higher number of generations could lead to improved results. We decided to introduce the defined stopping criteria due to the high computational demands of the framework, stemming from the numerous optimization problems. Despite this, improved performances have been obtained w.r.t. non linear optimization approach. The bilevel optimization resulted in an optimized robot design with a height of $\approx 1.49 \meter$, slightly taller than the design from nonlinear optimization ($\approx 1.47 \meter$) despite considering a narrower range of object heights. This difference can be attributed to the influence of collaborating with diverse human agents, which informed the ergonomics requirements in the bilevel optimization. In terms of motor optimization, the bilevel approach assigned L motors to the shoulder roll for load lifting and allocated greater strength to the ankle pitch and hip yaw motors for efficient weight-bearing.\looseness=-1
\subsection{Static Case}
\label{sec:validation:staticCase}
\begin{figure}[!t]
\centering
\includegraphics[trim=3.0cm 0.0cm 3.0cm 2.0cm, clip=true, width=\columnwidth]{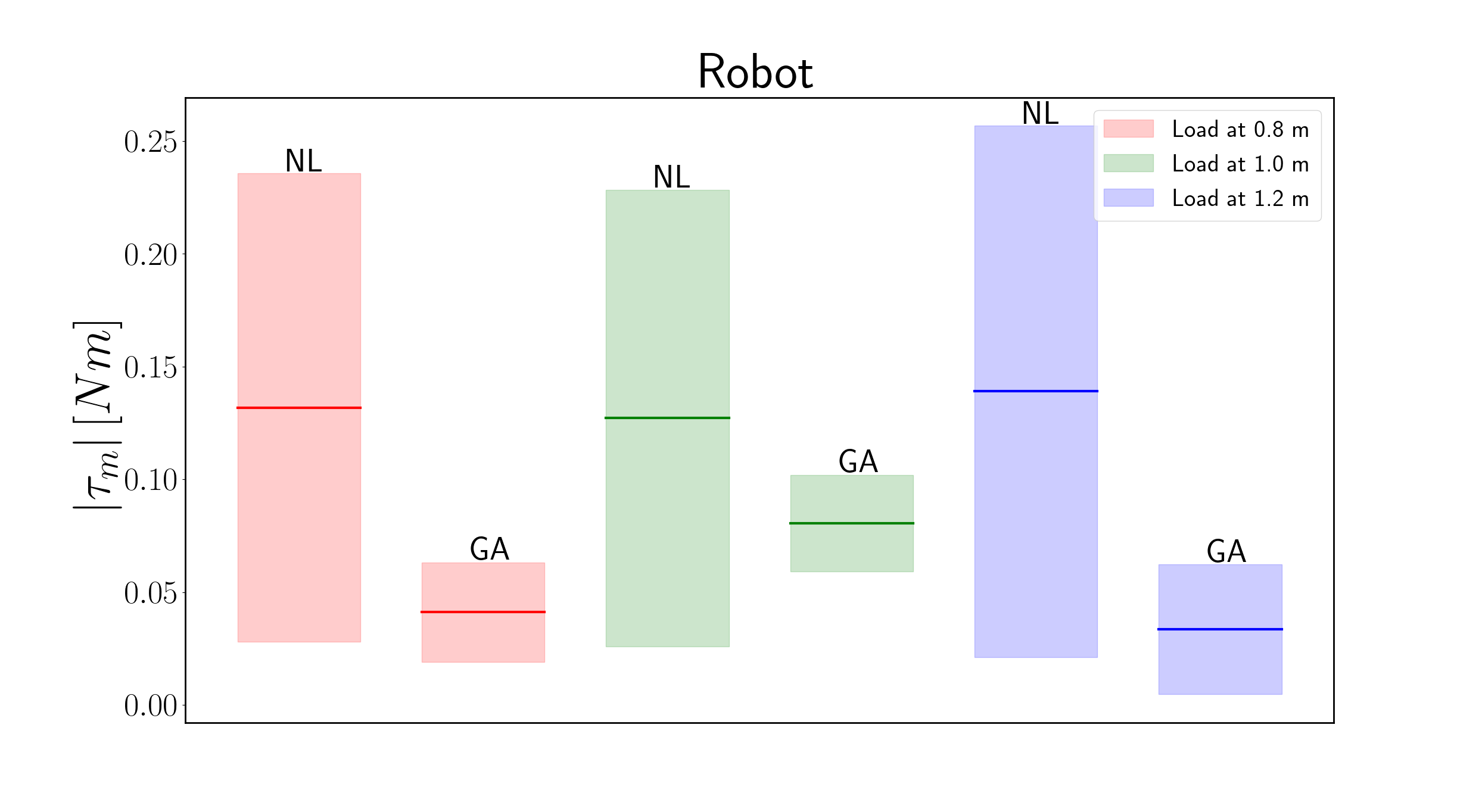}
\caption{The mean and variance of the absolute value of the robot motor torque for three different load heights for the nonlinear optimization (NL) output design and the Genetic Algorithm (GA) output design.} 
\label{fig:robot_torques}
\end{figure}

The static validation compares the output of the proposed approach with the output of nonlinear optimization performed as per \cite{sartore2022optimization}. The task taken into account is a payload lifting task performed with a human, $1.66 \text{ } \meter$ tall, that was not considered in the optimization process of both designs. The load taken into account in the static validation weighs  $5 \text{ } \kilogram$. The optimal static configuration for each load height per robot model has been computed using the same optimization problem as Eq. \eqref{eq:optimization}.  The robot motor torques for both outputs are depicted in Fig. \ref{fig:robot_torques}. As can be noticed, the output of the genetic algorithms shows lower torque for all the heights. Indeed the robot energy expenditure is decreased overall by $43\%$ w.r.t. the nonlinear optimization output.
Table \ref{tab:huamn-torque} compares human back torque for each load height during collaboration with robot designs optimized via genetic algorithm and non linear optimization. The genetic algorithm output reduces back torque for the more challenging heights of $0.8\meter$ and $1.2\meter$ which requires bending and elongation. The most significant improvement observed at the  $L5S1$ joint, which is one of the most critical one for the human \cite{lavender2003effects}. Indeed, the genetic algorithm output is able to reduce the $L5S1$ human torque by approximately 30\% for the $1.2 \meter$ load height.
\begin{figure}[!t]
\centering
    \includegraphics[trim=3.0cm 0.0cm 3.0cm 2.0cm, clip=true, width=\columnwidth]{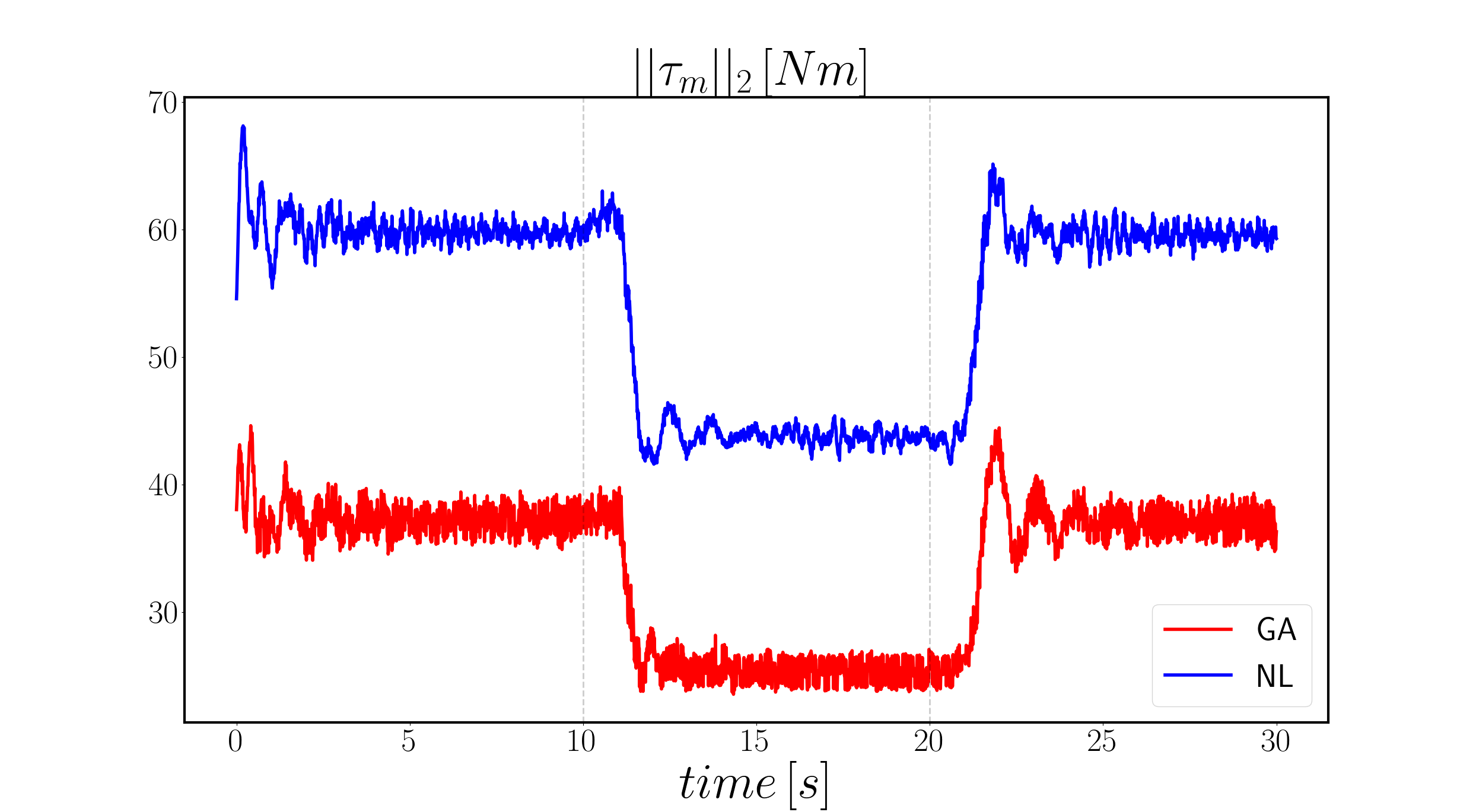}
    \caption{Torque norm over time for the design obtained via genetic algorithm (GA) and via nonlinear optimization (NL) performing a payload lifting task.}
    \label{fig:simulationOutput}
\end{figure}
\begin{table}
\begin{center}
\caption{Human back-torque for each load height during collaboration with Genetic Algorithm robot (GA) and NonLinear Optimized robot (NL),including the difference between the two. Blue highlights GA's lower back-torque, while red indicates NL's lower back torque.
}
\label{tab:huamn-torque}
\begin{tabular}{|c|c|c|c|}
\hline
\backslashbox{\textbf{Joint} [Nm]}{\textbf{Height} [m]}
&\makebox[3em]{\textbf{0.8}}&\makebox[3em]{\textbf{1.0}}&\makebox[3em]{\textbf{1.2}}\\ \hline
GA $\boldsymbol{T9T8}$& -47.76 & -33.81 & -24.86 \\ \hline
NL $\boldsymbol{T9T8}$& -49.00 & -32.46 & -25.24 \\ \hline
$\left(|GA| - |NL|\right)$ $\boldsymbol{T9T8}$ &\cellcolor{blue!25} -1.24 &\cellcolor{red!25} 1.35 &\cellcolor{blue!25} -0.38 \\ \hline
GA $\boldsymbol{L1T12}$& -51.336 & -31.90 & -19.06 \\ \hline
NL $\boldsymbol{L1T12}$& -52.22 & -30.61 & -20.35 \\ \hline
$\left(|GA| - |NL|\right)$  $\boldsymbol{L1T12}$&\cellcolor{blue!25} -0.884 &\cellcolor{red!25} 0.99 &\cellcolor{blue!25} 1.29 \\ \hline
GA $\boldsymbol{L4L3}$& -49.55 & -28.36 & -12.19 \\ \hline     
NL $\boldsymbol{L4L3}$& -50.22 & -27.42 & -14.17 \\ \hline 
$\left(|GA| - |NL|\right)$  $\boldsymbol{L4L3}$&\cellcolor{blue!25} -0.67 &\cellcolor{red!25} 0.94 &\cellcolor{blue!25} -1.98 \\ \hline 
GA $\boldsymbol{L5S1}$& -45.84 & -22.93 & -5.42 \\ \hline     
NL $\boldsymbol{L5S1}$& -46.41 & -22.46 & -7.82 \\ \hline 
$\left(|GA| - |NL|\right)$  $\boldsymbol{L5S1}$&\cellcolor{blue!25} -0.51 &\cellcolor{red!25} 0.5 &\cellcolor{blue!25} -2.4 \\ \hline 
\end{tabular}
\end{center}
\end{table}
\subsection{Dynamic Case}
\label{sec:validation:simulation}
The output of the proposed approach has been compared to the design obtained with the nonlinear optimization as per \cite{sartore2022optimization} in simulation while performing a single agent payload lifting task. \emph{Gazebo} \cite{koenig2004design} has been used as a simulator. The robots were torque-controlled and the momentum-based controller of \cite{rapetti2023} has been used to compute the input torque. The robot starts from an initial configuration, holding a load at $0.55 \meter$, then at $t=10s$ the load is lifted by the robot to the height of $0.75\meter$, and the robot balances in such a configuration. Finally, at $t=30s$ the robot goes back to the initial configuration and keeps it for $10 s$. We decided to consider a load of $1 \kilogram$ to evaluate the output design in task settings not considered in the optimization process. The torque norm over time is depicted in Fig. \ref{fig:simulationOutput}. As can be noticed, the genetic algorithm output shows lower torque both during the static phase (hence while balancing) and also during the lifting and lower movements. More into detail, the overall torque was decreased by $38.77\%$, leading to an improvement in the robot ergonomy.  The GA design torque exhibits high-frequency oscillations, attributed to the usage of the same set of control gains for both models. However, further controller tuning can improve this behavior.
\section{Conclusions}
\label{sec:conclusions}
This paper presents a method for optimizing humanoid robot motor characteristics and links inertial parameters for physical human-robot collaborations using bilevel optimization leveraging genetic algorithms and nonlinear optimization. The optimization encompasses a payload lifting task with different humans and loads. 
The results obtained consider whole body optimization and exploit the interplay between the different robot parts in influencing human ergonomics. Nevertheless, refinement and further tests will be conducted, evaluating each robot subpart individually, to port the results into the real ergoCub2 hardware.
Despite the results obtained, we considered only static conditions in the optimization process. 
Anyhow the proposed dynamic parametrization allows in the future to consider the time evolution of the system in the optimization process. In addition, such a formulation is aligned with the definition of the whole body controller of \cite{rapetti2023}, thus allowing co-design with respect to human metrics. In future work, we intend to address the aforementioned points  enhancing the computational efficiency of the framework, accounting for the time evolution of the system, and optimizing control and hardware parameters together to improve human ergonomics.
\bibliographystyle{IEEEtran}
\bibliography{root}

\end{document}